\documentclass[10pt,twocolumn,letterpaper]{article}

\usepackage[pagenumbers]{cvpr} 

\definecolor{cvprblue}{rgb}{0.21,0.49,0.74}
\usepackage[pagebackref,breaklinks,colorlinks,allcolors=cvprblue]{hyperref}
\usepackage{algorithm}
\usepackage{algpseudocode}
\algrenewcommand\algorithmicrequire{\textbf{Input:}}
\algrenewcommand\algorithmicensure{\textbf{Output:}}
\newcommand\blfootnote[1]{%
  \begingroup
  \renewcommand\thefootnote{}\footnote{#1}%
  \addtocounter{footnote}{-1}%
  \endgroup
}

\usepackage[table]{xcolor} 
\usepackage{makecell}      
\usepackage{enumitem}
\usepackage{bbding}
\usepackage{graphicx}
\usepackage{amsmath}
\usepackage{amssymb}
\usepackage{booktabs}
\usepackage{pifont}
\usepackage{color}
\usepackage{multirow}
\usepackage{overpic}
\usepackage{arydshln}
\usepackage{makecell}
\usepackage{times}
\usepackage{epsfig}
\usepackage[T1]{fontenc}
\usepackage{hhline}
\usepackage{xcolor}
\usepackage{pifont}
\usepackage{enumitem}
\usepackage{colortbl}
\usepackage{verbatim}
\usepackage{bm} 
\usepackage{wrapfig}
\definecolor{mygray}{gray}{.92}

\usepackage{amsmath, amssymb}    
\usepackage{lipsum} 

\usepackage{mathtools}

\usepackage{amsmath,amsfonts,bm}

\def\eqref#1{equation~\ref{#1}}

\def\1{\bm{1}}

\def\va{{\bm{a}}}

\def\vc{{\bm{c}}}

\def\vs{{\bm{s}}}

\def\vv{{\bm{v}}}

\def\vx{{\bm{x}}}

\DeclareMathAlphabet{\mathsfit}{\encodingdefault}{\sfdefault}{m}{sl}
\SetMathAlphabet{\mathsfit}{bold}{\encodingdefault}{\sfdefault}{bx}{n}

\usepackage{hyperref}
\usepackage{url}
\usepackage[capitalize,noabbrev]{cleveref}
\newcommand{\ourmethodabbr}{VGPO} %
\newcommand{\ourmethod}{Value-Anchored Group Policy Optimization}

\def\paperID{7705} 
\def\confName{CVPR}
\def\confYear{2026}

\title{Anchoring Values in Temporal and Group Dimensions \\for Flow Matching Model Alignment}
\author{Yawen Shao$^{1}$, Jie Xiao$^{1,2}$, Kai Zhu$^{1,2,\dagger}$, Yu Liu$^{2}$, Wei Zhai$^{1,\dagger}$, Yang Cao$^{1}$, Zheng-Jun Zha$^{1}$\\
{$^{1}$~University of Science and Technology of China}\qquad
{$^{2}$~Tongyi Lab} \\
\small{\texttt{shaoyawen@mail.ustc.edu.cn}}
}

\begin{document}
\maketitle
\begin{abstract}
\blfootnote{$\dagger$Corresponding Author.}Group Relative Policy Optimization (GRPO) has proven highly effective in enhancing the alignment capabilities of Large Language Models (LLMs). However, current adaptations of GRPO for the flow matching-based image generation neglect a foundational conflict between its core principles and the distinct dynamics of the visual synthesis process. This mismatch leads to two key limitations: (i) Uniformly applying a sparse terminal reward across all timesteps impairs temporal credit assignment, ignoring the differing criticality of generation phases from early structure formation to late-stage tuning. (ii) Exclusive reliance on relative, intra-group rewards causes the optimization signal to fade as training converges, leading to the optimization stagnation when reward diversity is entirely depleted. To address these limitations, we propose Value-Anchored Group Policy Optimization (VGPO), a framework that redefines value estimation across both temporal and group dimensions. Specifically, VGPO transforms the sparse terminal reward into dense, process-aware value estimates, enabling precise credit assignment by modeling the expected cumulative reward at each generative stage. Furthermore, VGPO replaces standard group normalization with a novel process enhanced by absolute values to maintain a stable optimization signal even as reward diversity declines. Extensive experiments on three benchmarks demonstrate that VGPO achieves state-of-the-art image quality while simultaneously improving task-specific accuracy, effectively mitigating reward hacking. Project webpage: \href{https://yawen-shao.github.io/VGPO/}{https://yawen-shao.github.io/VGPO/}.
\end{abstract}    
\section{Introduction}
\label{sec:intro}
\begin{figure*}[t]
\centering  
    \begin{overpic}[width=1\linewidth]{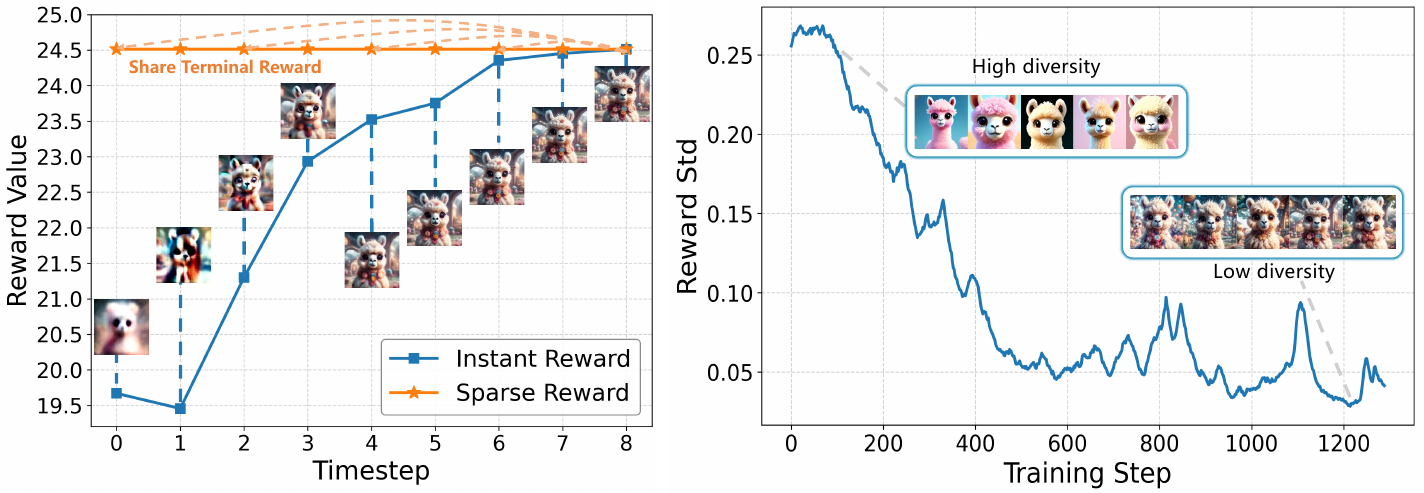}  
    \end{overpic} 
    \caption{\textbf{Motivation.} (Left) \textbf{Sparse vs. Instant Reward Signals During Generation.} The sparse terminal reward remains constant, failing to provide varying values for intermediate steps. (Right) \textbf{Diminishing Reward Std as Policy Converges.} Due to the reliance on reward diversity, the reward std declines as the training process advances, potentially leading to optimization stagnation.}
   
    \label{fig:motivation}
\end{figure*}

The evolution of aligning Large Language Models (LLMs)~\cite{guo2025deepseek,openai} with human intent has recently entered a new phase~\cite{ouyang2022training}. Initially dominated by supervised instruction tuning, the field is now increasingly leveraging the principles of reinforcement learning (RL)~\cite{rl} to achieve more sophisticated behavioral control. This transition is marked by recent advances such as PPO~\cite{ppo}, DPO~\cite{dpo} and GRPO~\cite{grpo}, which represent a departure from training on static datasets toward interaction-based optimization of policies driven by human preference signals.

Reinforcement learning for flow-based generative models remains comparatively underexplored, in contrast to diffusion-based generative models~\cite{ddpo,d-dpo}. Most efforts~\cite{liu2025improving,chen2025skyreels} directly transplant paradigms developed for large language models into the flow matching setting with minimal adaptation, neglecting the distinctive characteristics of generative process dynamics. For instance, Flow-GRPO~\cite{flow-grpo} and DanceGRPO~\cite{xue2025dancegrpo} propose to directly apply the advanced GRPO~\cite{grpo} algorithm to state-of-the-art text-to-image flow matching models~\cite{sd3.5,flux} through exploring the action space using SDE sampling methods~\cite{song2020score}. 

However, these methods tend to overlook the potential mismatch between the assumptions of GRPO and the dynamics of the flow matching environment.
First, GRPO assumes that the action values are uniform across all intermediate steps. 
In the environment of flow matching models, the progressive transformation of Gaussian noise into a high-quality image introduces intermediate actions of varying values.
By distributing a uniform, terminal reward across all denoising steps, these methods ignore the differential impact of each step in the generation process, failing to distinguish between critical early stages for structure formation and later stages for fine-grained detail refinement.
%
This indiscriminate credit allocation results in misleading optimization signals that impair sample efficiency and hinder effective learning~\cite{yang2024dense,qu2025survey,cui2025entropy}. 
%
As shown in Fig.~\ref{fig:motivation} (Left), this sequential process enables the evaluation of intermediate action values, offering a more granular assessment than previous methods reliant solely on a share terminal reward.
Second, GRPO~\cite{grpo} and its related methods~\cite{dapo} fundamentally leverage the diversity within final rewards to guide their optimization process.
The optimization signal is only derived from the relative advantage, which depends on reward variance within a group. 
Nevertheless, our empirical result (Fig.~\ref{fig:motivation} (Right)) indicates that the diversity, which serves as the driving force for optimization, progressively diminishes as the optimization process advances.
This can cause the optimization process to stagnate, when the model exclusively generates uniformly low/high-reward results within a rollout group.
This vulnerability to stagnation is particularly acute in visual generation tasks compared to large language models, as models can more easily converge to a single aesthetic or stylistic mode~\cite{lee2023aligning}.

To address these limitations, we propose \textbf{V}alue-Anchored \textbf{G}roup \textbf{P}olicy \textbf{O}ptimization (\textbf{\ourmethodabbr}) framework built on two key components. 
First, we introduce the Temporal Cumulative Reward Mechanism (TCRM), which leverages Monte-Carlo estimation over the sampling trajectory to assess the value of intermediate actions.
Specifically, we introduce the definition of an instant reward for a given state-action pair ($\vs_t$, $\va_t$), which subsequently allows to approximate the ground-truth intermediate action value using the available sampled trajectories.
Second, to counteract the adverse effects of diminishing reward diversity, we propose the Adaptive Dual Advantage Estimation  (ADAE). This replaces standard normalization with a novel process enhanced by absolute metrics for advantage computation. Critically, we can prove that ADAE automatically switches to optimizing absolute values when reward diversity is fully depleted.
Extensive experiments on compositional image generation, visual text rendering and human preference alignment tasks demonstrate that \ourmethodabbr\ enhances task-specific accuracy while significantly improves image quality and diversity.

The contributions of this paper are as follows:
\begin{itemize}
\item We identify a fundamental mismatch between the core assumptions of GRPO and the dynamics of flow-based generation, leading to two critical limitations: misalignment between process exploration and outcome reward, and reliance on reward diversity.
\item We propose \ourmethodabbr, a framework built upon the synergistic action of the temporal cumulative reward mechanism for process-aware value estimation and the adaptive dual advantage estimation for stable advantage computation.
\item We conduct comprehensive experiments on three benchmarks, demonstrating that VGPO achieves state-of-the-art performance by attaining higher alignment accuracy, promoting more efficient exploration, and mitigating reward hacking.
\end{itemize}
\section{Related work}
\label{sec:related work}
\textbf{RL for Diffusion Models.}
Due to the significant effectiveness of RL in enhancing the reasoning capabilities of large language models (LLMs)~\cite{openai,grpo}, its application to diffusion models has become a rapidly developing research direction. 
Early works~\cite{d3po,ddpo,dpok,lee2023aligning} draw inspiration from classical policy gradient algorithms like Proximal Policy Optimization (PPO) ~\cite{ppo} to align pretrained T2I models with human preferences. Subsequently, Diffusion-DPO~\cite{d-dpo} and its variants~\cite{spo,raft,yuan2024selfplayfinetuningdiffusionmodels,yuan2024self} integrate Direct Preference Optimization (DPO)~\cite{dpo} into T2I generation to enable direct learning from preference data.
%
Recent works~\cite{mixgrpo,tempflowgrpo,branchgrpo,flow-cps,prefgrpo,diffusiontrajectory,zheng2025diffusionnftonlinediffusionreinforcement} begin to explore the potential of online RL in advancing flow matching generative models. In particular, Flow-GRPO~\cite{flow-grpo} and DanceGRPO~\cite{xue2025dancegrpo} are the first to incorporate advanced Group Relative Policy Optimization (GRPO)~\cite{grpo} into flow-matching models by converting ODE sampling into equivalent SDE. 
However, directly transplanting GRPO into the flow-matching setting fails to account for the mismatch between the algorithm’s core assumptions and the intrinsic dynamics of flow matching.  
Our \ourmethodabbr\ addresses this by introducing the Temporal Cumulative Reward Mechanism (TCRM) to establish a process-aware reward structure and the Adaptive Dual Advantage Estimation (ADAE) to prevent policy collapse by maintaining a stable optimization.

\textbf{Dense Process Rewards.}
The challenge of credit assignment with sparse terminal rewards has driven the adoption of dense process rewards, which have proven effective in areas such as the inference-time scaling of LLMs~\cite{llm1,llm2,prime}.
TPO~\cite{liao2024tpo} extracts more fine-grained process rewards by ranking entire reasoning trajectories and adaptively assigning credit to the critical intermediate steps.
Recent efforts to apply dense process rewards in diffusion models have explored two main paradigms. The first involves building explicit process reward models (PRMs), such as in SPO~\cite{spo}, which trains a model to evaluate intermediate steps. However, this method is often hampered by high annotation costs and the challenge of training on noisy images.
The second paradigm infers process signals from terminal rewards. For example, DenseReward~\cite{yang2024dense} breaks temporal symmetry in DPO-style objectives by introducing temporal discounting. 
%
However, prior credit assignment methods are highly sample-inefficient, requiring full trajectory rollouts for single-step evaluation, and myopic, attributing terminal rewards to single actions without considering long-term value. 
Our \ourmethodabbr\ resolves these limitations by efficiently estimating long-term cumulative values from one-step ODE sampling and in turn using them to re-weight timesteps, assigning greater importance to critical decisions in the generation process.
\section{Method}
\subsection{Preliminaries}
\textbf{Flow Matching.} The Rectified Flow~\cite{rectifiedflow,flow-matching} framework has emerged as a foundational technique for generative modeling, underpinning recent advances in both image and video generation. Central to this framework is the construction of a linear trajectory that connects a data sample  $\vx_0\sim{X}_{0} $ with a noise sample $\vx_1\sim{X}_{1}$. A noisy latent $\vx_{t}$ is defined as:
\begin{equation}
\label{Eq:1}
\vx_{t} = (1-t) \vx_{0}+t \vx_{1}
\end{equation}
By training the model to predict the velocity $\vv$, the Flow Matching objective can be formulated as:
\begin{equation}
\label{Eq:2}
\boldsymbol{L}(\theta)=\mathbb{E}_{t, \vx_{0}, \vx_{1} }\left\|\vv-\vv_{\theta}\left(\vx_{t}, t\right)\right\|^{2}
\end{equation}
where the target velocity field is $\vv = \vx_{1}-\vx_{0}$.

\textbf{GRPO.} Group Relative Policy Optimization (GRPO)~\cite{grpo} is a reinforcement learning method that leverages the average reward of multiple sampled outputs as a dynamic baseline for advantage estimation. This principle was recently adapted for generative models in Flow-GRPO~\cite{flow-grpo}, which applies GRPO to improve the performance of state-of-the-art flow matching models~\cite{wan,sd3.5,flux}. 
The underlying framework for this approach, following prior work on diffusion models~\cite{ddpo}, is to cast the iterative generation process as a Markov Decision Process (MDP), formulated as:
\begin{equation} 
\label{Eq:3}
\begin{split}  
    & \vs_{t} \triangleq\left(\vc, t, \vx_{t}\right) \quad \pi\left(\va_{t} \mid \vs_{t}\right) \triangleq p_{\theta}\left(\vx_{t-1} \mid \vx_{t}, \vc\right) \\
    & \va_{t} \triangleq \vx_{t-1} \qquad P\left(\vs_{t+1} \mid \vs_{t}, \va_{t}\right) \triangleq\left(\delta_{\vc}, \delta_{t-1}, \delta_{\vx_{t-1}}\right) \\
    & \rho_{0}\left(\vs_{0}\right) \triangleq\left(p(\vc), \delta_{T}, \mathcal{N}\left(\boldsymbol{0}, \mathbf{I}\right)\right) \quad R\left(\vs_{t}, \va_{t}\right) \triangleq r\left(\vx_{0}, \vc\right) 
\end{split}
\end{equation}
where at each timestep $t$, the agent observes a state $\vs_{t}$, takes an action $\va_{t}$, receives a reward $R\left(\vs_{t}, \va_{t}\right)$, and transitions to a new state $\vs_{t+1} \sim  P\left(\vs_{t+1} \mid \vs_{t}, \va_{t}\right)$. The agent acts according to a policy $\pi\left(\va_{t} \mid \vs_{t}\right)$ and $\rho_{0}\left(\vs_{0}\right)$ represents the initial-state distribution.

Given a prompt $\vc$, the flow model $p_{\theta}$ samples a group of $G$ individual images $\left\{\vx_{0}^{i}\right\}_{i=1}^{G}$ and the corresponding reverse-time trajectories $\left\{\vx_{T}^{i},\vx_{T-1}^{i},\cdots,\vx_{0}^{i}\right\}_{i=1}^{G}$. Then, the advantage of the $i$-th image is calculated by normalizing the group-level rewards as follows:
\begin{equation}
\label{Eq:4}
\hat{A}_{t}^{i}=\frac{R\left(\vx_{0}^{i}, \vc\right)-\operatorname{mean}\left(\left\{R\left(\vx_{0}^{i}, \vc\right)\right\}_{i=1}^{G}\right)}{\operatorname{std}\left(\left\{R\left(\vx_{0}^{i}, \vc\right)\right\}_{i=1}^{G}\right)}
\end{equation}
Flow-GRPO optimizes the policy model by maximizing the following objective:
\begin{equation}
\label{Eq:5_corrected}
\resizebox{1.0\columnwidth}{!}{ 
$\begin{gathered}
\mathcal{J}_{\text {Flow-GRPO}}(\theta) = \mathbb{E}_{\vc \sim \mathcal{C},\left\{\vx^{i}\right\}_{i=1}^{G} \sim \pi_{\theta_{\text {old }}}(\cdot \mid \vc)} \Bigg[ \bigg( \frac{1}{G} \sum_{i=1}^{G} \frac{1}{T} \sum_{t=0}^{T-1} \min \Big( r_{t}^{i}(\theta) \hat{A}_{t}^{i}, \\
\operatorname{clip}(r_{t}^{i}(\theta), 1-\varepsilon, 1+\varepsilon) \hat{A}_{t}^{i} \Big) \bigg) - \beta D_{\mathrm{KL}}(\pi_{\theta} \| \pi_{\mathrm{ref}}) \Bigg]
\end{gathered}$
}
\end{equation}
where 
\begin{equation}
\label{Eq:6}
r_{t}^{i}(\theta)=\frac{p_{\theta}\left(\vx_{t-1}^{i} \mid \vx_{t}^{i}, \vc\right)}{p_{\theta_{\text {old }}}\left(\vx_{t-1}^{i} \mid \vx_{t}^{i}, \vc\right)}
\end{equation}
Flow matching models which utilize deterministic ODE-based sampling, inherently lack the stochasticity required for the probabilistic policy updates in GRPO. To address this, Flow-GRPO converts the deterministic ODE into an equivalent SDE that matches the original model’s marginal probability density function at all timesteps~\cite{song2020score,albergo2023stochastic,domingo2024adjoint}. The final update rule is formulated as:
\begin{equation}
\label{Eq:7}
\resizebox{1.0\columnwidth}{!}{
$
\vx_{t+\Delta t}=\vx_{t}+\left[\vv_{\theta}\left(\vx_{t}, t\right)+\frac{\sigma_{t}^{2}}{2 t}\left(\vx_{t} \\ +(1-t) \vv_{\theta}\left(\vx_{t}, t\right)\right)\right] \Delta t+\sigma_{t} \sqrt{\Delta t} \epsilon
$
}
\end{equation}
where $\sigma_{t}=a\sqrt{\frac{t}{1-t}}$ and $a$ is a scalar hyper-parameter that controls the noise level.
\begin{figure*}[t]
\centering  
    \begin{overpic}[width=1\linewidth]{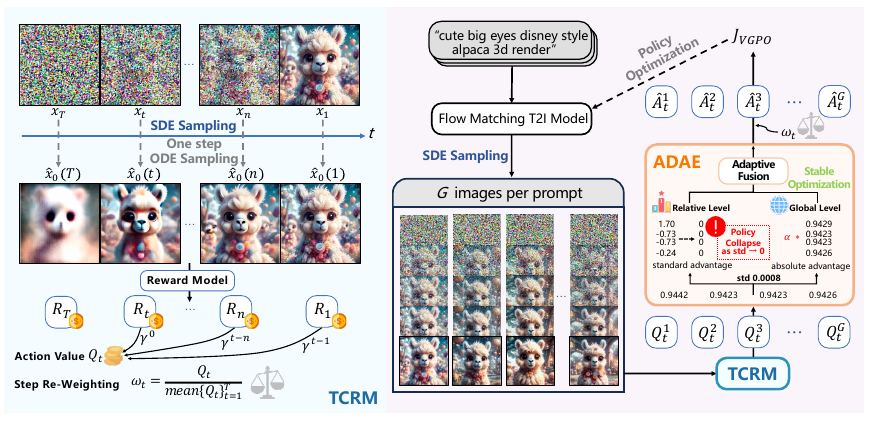}  
    \end{overpic} 
    \caption{\textbf{Method Overview.} First, to resolve faulty credit assignment,  Temporal Cumulative Reward Mechanism (TCRM) transforms sparse terminal rewards into dense, forward-looking process values, enabling a more granular, temporally-aware credit assignment. Second, to counteract policy collapse, Adaptive Dual Advantage Estimation (ADAE) replaces standard normalization with a novel process enhanced by absolute values for advantage computation, ensuring a persistent optimization that remains stable even when reward diversity diminishes.
}
    \label{fig:network}
\end{figure*}

\subsection{Motivation}
\textbf{Misalignment between Process Exploration and Outcome Reward.} The core limitation of applying GRPO to flow matching models is the temporal misalignment inherent in its objective function. This issue stems from coupling a time-dependent policy ratio, which represents the process exploration, with a time-independent outcome reward:
\begin{align}
\def\myfrac{\frac{p_{\theta}\left(\vx_{t-1}^{i} \mid \vx_{t}^{i}, \vc\right)}{p_{\theta_{\text {old }}}\left(\vx_{t-1}^{i} \mid \vx_{t}^{i}, \vc\right)}}
\mathcal{J} = \mathbb{E} \underset{
    \underbrace{\hphantom{\myfrac}}_{\mathclap{\text{process exploration}}} \cdot \underbrace{\hphantom{A}}_{\mathclap{\text{outcome reward}}}
}{
    \left[ \myfrac \cdot A(\vx_0) \right]
}
\end{align}
where the stepwise advantage remains constant by setting $A_t \equiv A(\vx_0)$ according to the final result. This formulation effectively distributes a uniform, sparse terminal reward across all timesteps, ignoring the differential impact of each action in the generative sequence. As illustrated in Fig.~\ref{fig:motivation} (Left), such indiscriminate credit assignment fails to capture the true value evolution as an image is progressively refined from noise. For instance, it may unduly penalize critical early-stage structural decisions or reward trivial late-stage refinements, resulting in misleading optimization signals that impair learning efficiency~\cite{guo2021efficient}. To resolve this misalignment, we propose a forward-looking temporal cumulative reward mechanism that aligns the reward signal with the exploration process, enabling more precise and efficient policy optimization.

\textbf{Reliance on Reward Diversity.} The GRPO framework leverages reward diversity within each generated group to derive its optimization signal, as its advantage function is normalized by the reward standard deviation (Eq.~\ref{Eq:4}). 
This driving force of optimization progressively diminishes as the optimization process advances, as shown in Fig.~\ref{fig:motivation} (Right). 
This dependency leads to policy stagnation whenever reward diversity is depleted, regardless of the absolute quality of the samples.
To address it, we propose adaptive dual advantage estimation, which ensures a persistent optimization gradient by decoupling the learning signal from reward variance, enabling continuous exploration towards higher-quality outcomes.

\subsection{\ourmethod}
In this section, we introduce \ourmethod\ (\ourmethodabbr), a novel framework (Fig.~\ref{fig:network}) that makes two core contributions: Temporal Cumulative Reward Mechanism (Sec.~\ref{3.3.1}) and Adaptive Dual Advantage Estimation (Sec.~\ref{3.3.2}).

\subsubsection{Temporal Cumulative Reward Mechanism} 
\label{3.3.1}
We introduce the Temporal Cumulative Reward Mechanism (TCRM) to transform sparse terminal rewards into a dense, forward-looking value signals for precise credit assignment. First, we define an instant reward for each state-action pair ($\vs_t$, $\va_t$), which subsequently enables the approximation of ground-truth intermediate action values using sampled trajectories. Second, we estimate the long-term cumulative value of each action $\va_t$ to overcome the myopia of greedy optimization, and utilize these values to dynamically re-weight the importance of each timestep in policy updates. The specifics of this mechanism are detailed below.

\textbf{Instant Reward.}
We formalize the generation process as a finite-time Markov Decision Process (MDP), characterized by the tuple $\left(\mathcal{S}, \mathcal{A}, \rho_{0},P, R\right)$. At each reverse-time step $t \in \{T, ..., 1\}$, the model is in a latent state $\vs_t = \vx_t \in \mathcal{S}$ and takes an action $\va_t \sim \pi(\cdot|\vs_t)$, which corresponds to the SDE exploration that transitions the state to $\vs_{t-1} = \vx_{t-1}$. A key challenge in this MDP is the absence of intermediate reward signals $R_t(\vs_t, \va_t)$. This issue is exacerbated in diffusion models, where the heavy noise in early-stage images makes direct evaluation unreliable and semantically meaningless via a process reward model~\cite{spo}. To avoid this problem, we conceptualize the flow model as a one-step generation model. Specifically, at each sampling step $t$, after taking action $\va_t$ in the state $\vs_t$ to reach $\vs_{t-1}$, we perform a one-step deterministic ODE sampling from $\vs_{t-1}$ to obtain a projected terminal state $\hat{\vx}_0$. The instant reward $R_t\left(\vs_t,\va_t\right)$ is defined as:
\begin{equation}
\label{Eq:9}
R_t\left(\vs_t, \va_t\right) = \mathrm{RM}\left(\hat{\vx}_0, \vc\right),\vs_{t-1} = f\left(\vs_t, \va_t\right)
\end{equation}
\begin{equation}
\label{Eq:10}
\hat{\vx}_0 = \vs_{t-1} - \tau_{t-1}\vv_{\theta}\left(\vs_{t-1}, \tau_{t-1}\right)
\end{equation}
where ${\vv}_{\theta}\left(\vs_{t-1},\tau_{t-1}\right)$ denotes velocity field predicted by the model and $\tau_t = \frac{t}{T}$, $\mathrm{RM}$ denotes reward model, $f$ denotes the SDE exploration (Eq.~\ref{Eq:7}).

\textbf{Long-term Cumulative Value.}
While the instant reward $R_t$ provides valuable per-step feedback, the policy that greedily optimizes for it would be myopic. Such a policy ignores the long-term consequences of an action, where a high immediate reward might steer the trajectory towards a suboptimal future. To instill long-term foresight into the policy, we estimate the action value function $Q^\pi(\vs_t, \va_t)$, which captures the expected cumulative discounted reward starting from action $\va_t$ in state $\vs_t$ and subsequently following policy $\pi$, formulated as:
\begin{equation}
\label{Eq:11}
Q^\pi\left(\vs_t, \va_t\right) = \mathbb{E}_{\pi} \left[ \sum_{k=0}^{t-1} \gamma^k R_{t-k} \mid \vs_t, \va_t \right]
\end{equation}
where $\gamma \in \left [ 0,1 \right ) $ is the discount factor. In practice, we leverages Monte-Carlo estimation~\cite{sutton1998reinforcement} over the sampling trajectory to assess the value $Q_t^i\left(\vs_t, \va_t\right)$ of intermediate actions.

While the action value $Q_t^i\left(\vs_t, \va_t\right)$ encapsulates the expected cumulative future reward, its absolute magnitude is discarded during standard advantage normalization.
This transformation to a relative-only signal obscures the intrinsic value of each timestep, preventing the model from recognizing which actions contributed more significantly to the overall outcome.
To recover this crucial information, we propose an explicit, value-driven weight $\omega_t^i$, which is designed to amplify the optimization signal for timesteps that lead to higher overall returns, dynamically assigning greater importance to more critical decisions within the generation process. It is formulated as:
\begin{equation}
\label{Eq:12}
\omega_t^i= \frac{Q_t^i\left(\vs_t, \va_t\right)}{\operatorname{mean}\left(\left\{Q_t^i\left(\vs_t, \va_t\right)\right\}_{t=1}^{T}\right)}
\end{equation}
\begin{algorithm*}[t]\caption{\ourmethodabbr\ Training Process}
\label{alg}
\begin{algorithmic}[1]
\Require Reward model $\mathrm{RM}$; Prompt dataset $\mathcal{C}$; Sampling steps $T$; Training steps $S$; Number of samples per prompt $G$; Temporal discount factor $\gamma$; Hyper-parameter $\alpha $
\Ensure Optimized model parameters $\theta$
\State Initial policy model $\pi_{\theta}$
\For {step $=1,\cdots,S$ }   
    \State Sample batch of prompts $C_b \sim \mathcal{C}$
    \State Update old policy model: $\pi_{\theta_{\text{old}}} \leftarrow \pi_{\theta}$
    \For{each prompt $\vc \in C_b$}
        \State Init the noise $\vx_{T} \sim \mathcal{N}(0, \mathbf{I})$
        \For{generate $i$-th image from $i = 1$ \textbf{to} $G$}
            \For{sampling timestep $t=T$ \textbf{to} $1$}
                \State Use SDE Sampling to get $\vx_{t-1}^{i}$ $\leftarrow$ Eq.~\ref{Eq:7}
                \State Use One-Step ODE Sampling to get $\hat{\vx}_{0}^{i}$ and instant reward  $R^i_t \leftarrow$ Eq.~\ref{Eq:9},~\ref{Eq:10}
            \EndFor
            \State Calculate long-term value $Q_{t}^{i} \leftarrow$ Eq.~\ref{Eq:11}
            \State Calculate value-driven weight $\omega_{t}^{i} \leftarrow$ Eq.~\ref{Eq:12}
            \State Calculate adaptive dual advantage: $\hat{A}_{t}^{i} \leftarrow $ Eq.~\ref{Eq:14}
        \EndFor
    \EndFor
    \State Update policy model via gradient ascent: $\theta \leftarrow \theta + \eta \nabla_{\theta} \mathcal{J}$
\EndFor
\end{algorithmic}
\end{algorithm*}
\subsubsection{Adaptive Dual Advantage Estimation} 
\label{3.3.2}
The advantage function $A^\pi(\vs_t, \va_t)$ quantifies the relative value of an action $\va_t$ compared to the expected policy behavior at state $\vs_t$.
To compute this advantage without the overhead of training a separate state value function $V^\pi(\vs_t)$, GRPO instead employs a sample-based estimation. Specifically, it generates $G$  distinct trajectories from a single prompt $\vc$ and computes the advantage relative to the mean of these sampled outcomes, formulated as:
\begin{equation}
\label{Eq:13}
\hat{A}_{t}^{i}\left(\vs_t, \va_t\right)=\frac{Q_t^i\left(\vs_t, \va_t\right)-\operatorname{mean}\left(\left\{Q_t^i\left(\vs_t, \va_t\right)\right\}_{i=1}^{G}\right)}{\operatorname{std}\left(\left\{Q_t^i\left(\vs_t, \va_t\right)\right\}_{i=1}^{G}\right)}
\end{equation}
However, the standard GRPO advantage function $\hat{A}_{t}$ is a purely relative measure, which introduces critical flaws that destabilize optimization:
(i) By applying identical optimization signals to sample groups of disparate absolute quality but similar relative structures, the advantage function stifles exploration for globally optimal strategies, effectively trapping the policy in local optima defined by relative gains.
(ii) In low-variance stage, normalization by a near-zero standard deviation (eg. std=0.0008 in Fig.~\ref{fig:network}) forges an illusory advantage by amplifying trivial reward gaps, driving reward hacking over genuine quality improvements.
(iii) During the late stages of policy convergence, the advantage signal collapses to zero as reward variance disappears, causing optimization to stagnate and risking policy collapse regardless of the samples' absolute quality.
To address these flaws, we propose the Adaptive Dual Advantage Estimation (ADAE), formulated as:
\begin{equation}
\label{Eq:14}
\hat{A}_{t}^{i}\left(\vs_t, \va_t\right)=\omega_t^i \cdot \frac{\left(1+\alpha \right) \cdot Q_t^i-\operatorname{mean}\left(\left\{Q_t^i\right\}_{i=1}^{G}\right)}{\operatorname{std}\left(\left\{Q_t^i\right\}_{i=1}^{G}\right)} 
\end{equation}
where $\alpha = k \cdot \mathrm{std}(\{Q^i_t\}^G_{i=1})$, with $k$ as a constant tuned per task reward. As reward diversity vanishes, the advantage signal smoothly transitions to being proportional to the absolute value $k \cdot Q^i_t$, preventing optimization from stagnating (see Appendix B for detailed derivation). $\omega_t^i$ is value-driven weight for step re-weighting (Eq.~\ref{Eq:12}). By adaptively merging the relative advantage dependent on reward diversity with robust global advantage, ADAE resolves the above flaws, achieving stable optimization and enabling higher-quality and more diverse generation. We present the complete \ourmethodabbr\ training strategy in Algorithm \ref{alg}.

\section{Experiments}
\subsection{Experimental Setup} 
Following Flow-GRPO, we evaluate our method on three distinct tasks: compositional image generation in GenEval~\cite{ghosh2023geneval} , visual text rendering~\cite{textdiffuser} in OCR~\cite{seedream} and human preference alignment in PickScore~\cite{pickscore}. For all tasks, the objective is to maximize the reward score while preserving overall image quality. We adopt SD-3.5~\cite{sd3.5} as the base model, consistent with the baseline. To demonstrate that our method effectively mitigates reward hacking, we assess performance from two fronts: (i) Task-specific accuracy on in-distribution test sets. (ii) General image quality on DrawBench~\cite{drawbench}. The latter is measured by a suite of metrics encompassing image quality (Aesthetic~\cite{aesthetics}, DeQA~\cite{deqa}) and preference score (ImageReward~\cite{imagereward}, PickScore~\cite{pickscore}, UnifiedReward~\cite{unifiedreward}), ensuring that improvements in task alignment do not compromise generative quality. Please see Appendix A for details.
\definecolor{greenx}{HTML}{107e4a}
\definecolor{bluex}{HTML}{4e9eef} 
\begin{table*}[t]
\caption{\textbf{Comparison Results} on Compositional Image Generation, Visual Text Rendering, and Human Preference Alignment benchmarks, evaluated by task performance, image quality, and preference score. ImgRwd: ImageReward; UniRwd: UnifiedReward.}
\vspace{-4mm}
\label{table:all}
\begin{center}
\renewcommand{\arraystretch}{1.2}
\renewcommand{\tabcolsep}{1 pt}
\resizebox{\linewidth}{!}{
\begin{tabular}{l ccc cc ccc}
\toprule
\textbf{Model} & \multicolumn{3}{c}{\textbf{Task Metric}} & \multicolumn{2}{c}{\textbf{Image Quality}} & \multicolumn{3}{c}{\textbf{Preference Score}} \\
&GenEval$\uparrow$ & OCR$\uparrow$ & PickScore$\uparrow$ & Aesthetic$\uparrow$ & DeQA$\uparrow$ & ImgRwd$\uparrow$ & PickScore$\uparrow$ & UniRwd$\uparrow$ \\
\midrule
SD3.5-M & $0.63$ & $0.59$ & $21.72$ & $5.39$	&$4.07$	&$0.87$	&$22.34$	&$3.07$ \\
\multicolumn{8}{c}{Compositional Image Generation} \\
\midrule
Flow-GRPO (w/o KL) & $0.95$& - & - & $4.93$	&$2.77$	&$0.44$	&21.16	&2.59  \\ 
\rowcolor{mygray}
\ourmethodabbr\ (w/o KL)  & $0.97\left(\textcolor{bluex}{+0.02}\right)$ &- & - & $5.23\left(\textcolor{bluex}{+0.3}\right)$ & $3.45\left(\textcolor{bluex}{+0.68}\right)$ & $0.94\left(\textcolor{bluex}{+0.50}\right)$ & $22.00\left(\textcolor{bluex}{+0.84}\right)$  &  $3.00\left(\textcolor{bluex}{+0.41}\right)$ \\ 
Flow-GRPO (w/ KL) & $0.95$ & - & -  &$5.25$	&$4.01$& $1.03$	&$22.37$ &$3.18$  \\ 
\rowcolor{mygray}  
\ourmethodabbr\ (w/ KL)  & $0.96\left(\textcolor{bluex}{+0.01}\right)$ & - & - & $5.41\left(\textcolor{bluex}{+0.16}\right)$ & $4.05\left(\textcolor{bluex}{+0.04}\right)$ & $1.09\left(\textcolor{bluex}{+0.06}\right)$ & $22.59\left(\textcolor{bluex}{+0.22}\right)$ & $3.23\left(\textcolor{bluex}{+0.05}\right)$   \\
\multicolumn{8}{c}{Visual Text Rendering} \\
\midrule
Flow-GRPO (w/o KL) & - & $0.93$ & - & $5.13$ &	$3.66$	& $0.58$ &	$21.79$	&$2.82$ \\
\rowcolor{mygray}
\ourmethodabbr\ (w/o KL)  & - & $0.95\left(\textcolor{bluex}{+0.02}\right)$ & - & $5.33\left(\textcolor{bluex}{+0.2}\right)$ & $3.98\left(\textcolor{bluex}{+0.32}\right)$ & $0.90\left(\textcolor{bluex}{+0.32}\right)$ & $22.17\left(\textcolor{bluex}{+0.38}\right)$ & $3.07\left(\textcolor{bluex}{+0.25}\right)$ \\
Flow-GRPO (w/ KL) & - & $0.92$ & - & $5.32$ &	$4.06$	& $0.95$ &	$22.44$	&$3.14$ \\
\rowcolor{mygray}
\ourmethodabbr\ (w/ KL)  & - &$0.94\left(\textcolor{bluex}{+0.02}\right)$ & - & $5.34\left(\textcolor{bluex}{+0.02}\right)$ & $4.08\left(\textcolor{bluex}{+0.02}\right)$ & $0.98\left(\textcolor{bluex}{+0.03}\right)$ & $22.44\left(\textcolor{bluex}{+0.0}\right)$ & $3.14\left(\textcolor{bluex}{+0.0}\right)$ \\
\multicolumn{8}{c}{Human Preference Alignment} \\
\midrule
Flow-GRPO (w/o KL) & - & - & $23.41$ & $6.15$	&$4.16$	&$1.24$	&$23.56$	&$3.33$  \\
\rowcolor{mygray}
\ourmethodabbr\ (w/o KL)  & - & - & $23.55\left(\textcolor{bluex}{+0.14}\right)$ & $5.97\left(\textcolor{greenx}{-0.18}\right)$ & $4.18\left(\textcolor{bluex}{+0.02}\right)$ & $1.28\left(\textcolor{bluex}{+0.04}\right)$ & $23.70\left(\textcolor{bluex}{+0.14}\right)$ &  $3.34\left(\textcolor{bluex}{+0.01}\right)$ \\
Flow-GRPO (w/ KL) & - & - & $23.31$ &$5.92$	&$4.22$ & $1.28$	&$23.53$	&$3.38$  \\
\rowcolor{mygray}  
\ourmethodabbr\ (w/ KL)  & - & - & $23.41\left(\textcolor{bluex}{+0.10}\right)$ & $5.90\left(\textcolor{greenx}{-0.02}\right)$ & $4.23\left(\textcolor{bluex}{+0.01}\right)$ & $1.32\left(\textcolor{bluex}{+0.04}\right)$ & $23.61\left(\textcolor{bluex}{+0.08}\right)$ &$3.39\left(\textcolor{bluex}{+0.01}\right)$   \\
\bottomrule
\end{tabular}
}
\end{center}
\end{table*}
\subsection{Main Results} 
\begin{figure} 
    
    \centering
    \begin{overpic}[width=0.9\linewidth]{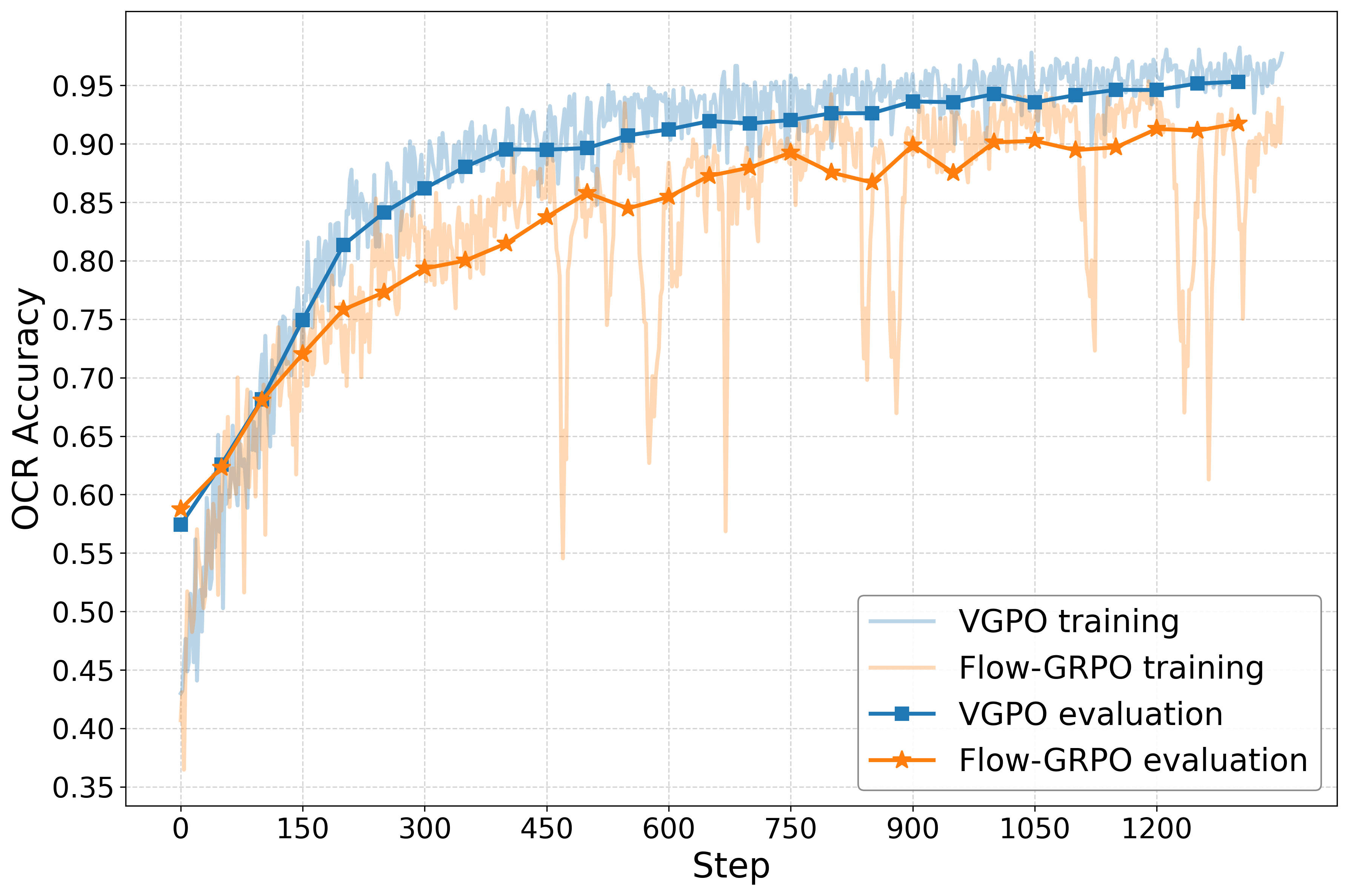}
    \end{overpic}
    \vspace{-4mm}
    \caption{\textbf{Learning Curves} with KL on OCR benchmark.}
    \label{fig:ocr_step} 
    \vspace{-10pt}
\end{figure}
\begin{figure*}[ht]
\centering  
    \begin{overpic}[width=1\linewidth]{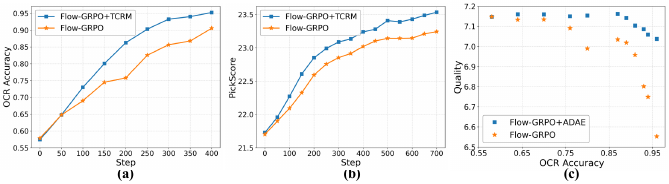} 
    \end{overpic} 
    \vspace{-6mm}
    \caption{\textbf{Ablation Analysis.} The impact of TCRM is evaluated on the (a) OCR and (b) PickScore benchmarks, while (c) assesses ADAE's contribution to image quality at an equivalent OCR accuracy. Quality is the average of the five image quality metrics.} 
    \label{fig:ablation}
\end{figure*}
\textbf{Quantitative Analysis.} Our quantitative evaluation, detailed in Tab.~\ref{table:all}, confirms the comprehensive superiority of \ourmethodabbr\ across three benchmarks. In the absence of KL regularization (w/o KL), \ourmethodabbr\ not only achieves steady improvements in task-specific metrics, but it enhances general image quality and preference score, significantly mitigating the reward hacking issue compared to Flow-GRPO.
In the compositional generation task without KL regularization (w/o KL), \ourmethodabbr\ improves the GenEval score by $0.02$ (from $0.95$ to $0.97$) while simultaneously boosting the average score across five quality and preference metrics by $9\%$. This pattern of dual improvement is also evident in visual text rendering, where OCR accuracy rises from $0.93$ to $0.95$ alongside substantial enhancements in image quality. We note that the minor drop in the aesthetic score during the human preference alignment task is not a quality collapse, but an expected consequence of the inherent coupling between the target reward and the evaluation metric~\cite{flow-grpo}.
Furthermore, this superiority persists under KL regularization. As depicted in Fig.~\ref{fig:ocr_step}, in addition to accelerated convergence(only $650$ training steps to match the peak performance of Flow-GRPO), \ourmethodabbr\ (w/ KL) exhibits improved training stability, culminating in a higher final accuracy.

 \begin{table}
    \caption{\textbf{Ablation Study} of main components.}
    \label{table:ablation}
    \vspace{-4mm}
    \begin{center} 
    \renewcommand{\arraystretch}{1.2}
    \renewcommand{\tabcolsep}{1pt}
    \small
    \resizebox{\linewidth}{!}{
    \begin{tabular}{cccccccc}
    \toprule
    TCRM & ADAE & OCR & Aes & DeQA & ImgRwd & PickScore & UniRwd\\
    \midrule
        & &$0.93$ & $5.13$ &	$3.66$	& $0.58$ &	$21.79$	&$2.82$\\
    \checkmark           &  & $0.94$	&$5.10$	&$3.86$	&$0.73$ &$21.98$	&$2.92$    \\
     &    \checkmark          & $0.94$ &	$5.21$	&$3.88$&	$0.86$	&$22.27$	& $3.02$\\
    \rowcolor{mygray}
    \checkmark & \checkmark   &  $0.95$ & $5.33$ & $3.98$ & $0.90$ & $22.17$ & $3.07$ \\
    \bottomrule
    
    \end{tabular}
     }
    \end{center}
    \label{tab:ablation_study}
\vspace{-5mm}
\end{table}
\begin{figure*}[t]
\centering  
    \begin{overpic}[width=0.94\linewidth]{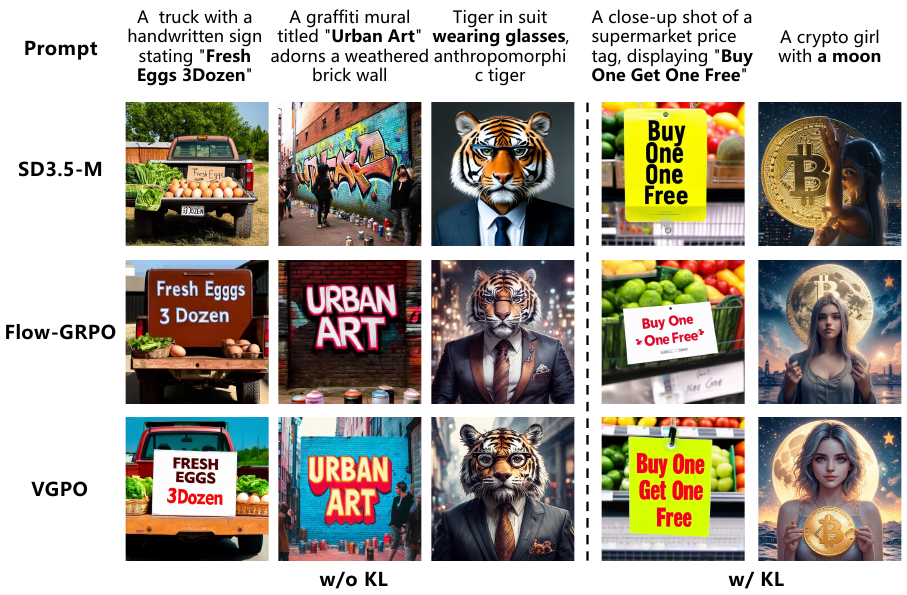}  
    \end{overpic} 
    \vspace{-3mm}
    \caption{\textbf{Qualitative Comparison.} \ourmethodabbr\ achieves superior performance in task accuracy, image quality and fine-grained details.}
    \label{fig:results}
    \vspace{-2mm}
\end{figure*}

\textbf{Qualitative Analysis.} The quantitative findings are further corroborated by our qualitative analysis, with representative visualizations presented in Fig.~\ref{fig:results}. For the visual text rendering task, the first and fourth columns highlight \ourmethodabbr's superior text accuracy. Notably, the second column reveals that \ourmethodabbr\ maintains strong visual diversity even after successfully rendering text, effectively resisting the tendency to overfit the reward and collapse into a single stylistic mode.
 In the human preference alignment task, examples in the third and fifth columns showcase \ourmethodabbr's exceptional capability in rendering fine-grained details and complex textures, producing images with heightened realism and visual fidelity. See Appendix C for per-category performance on the GenEval benchmark, and Appendix E for more visualizations.

\subsection{Ablation Analysis}
We conducted ablation studies to investigate the individual contributions of our two core components: TCRM and ADAE. Using the OCR task (w/o KL) as a case study, Tab.~\ref{table:ablation} shows that both components independently improve task accuracy and enhance overall image quality. TCRM's primary contribution is accelerating convergence, as shown in Fig.~\ref{fig:ablation} (a)(b). By providing a dense and granular optimization signal at each step, TCRM guides the model more efficiently towards the optimal solution, thereby reducing the sample inefficiency associated with sparse rewards. This allows the model to reach target performance in fewer training steps.
 In contrast, ADAE ensures training stability. It provides a robust optimization signal that prevents policy stagnation, and this sustained learning gradient translates directly into superior output quality. Fig.~\ref{fig:ablation} (c) confirms this, demonstrating that introducing ADAE leads to significant improvements in image quality and preference scores while maintaining an equivalent level of task performance. Crucially, this holistic enhancement is achieved without sacrificing visual quality, confirming that ADAE effectively mitigates reward hacking rather than narrowly overfitting to the reward signal.
 Ultimately, these components form a solution that holistically addresses inefficient credit assignment and policy stagnation.
\section{Conclusion}
In this paper, we observe that directly applying GRPO frameworks to flow matching models introduces two critical limitations: a misalignment between the exploration process and the final reward outcome, caused by the uniform application of sparse terminal rewards, and reliance on reward diversity renders it vulnerable to optimization stagnation as this diversity decreases. To address these problems, we propose Value-Anchored Group Policy Optimization (VGPO). VGPO facilitates granular credit assignment by transforming sparse terminal rewards into dense, forward-looking process values. Concurrently, it incorporates absolute values into the advantage computation to maintain a persistent optimization signal. Extensive experiments on three benchmarks demonstrate that VGPO achieves significant improvements in both task-specific accuracy and general image quality, while mitigating reward hacking.

\noindent\textbf{Limitation.} While our method's per-step reward calculation enhances generation quality, it introduces a manageable computational overhead. This cost can be significantly mitigated by using sglang to deploy the reward service, enabling synchronous computation of rewards for all steps. We contend this is a worthwhile trade-off, where a modest increase in computation time yields substantial improvements in final output quality and alignment. (See Appendix D for details.)
{
    \small
    \bibliographystyle{ieeenat_fullname}
    \bibliography{main}
}


\end{document}